\documentclass[11pt]{article}
\usepackage[final]{coling}
\usepackage{multirow}
\usepackage{times}
\usepackage{latexsym}
\usepackage{xurl}
\usepackage{enumitem}
\usepackage[T1]{fontenc}
\usepackage[utf8]{inputenc}
\usepackage{footnote}
\usepackage{microtype}
\usepackage{todonotes}
\usepackage{soul}
\usepackage{array}
\usepackage{inconsolata}
\usepackage{graphicx}
\usepackage{lipsum}
\usepackage{multicol}
\usepackage{booktabs}
\usepackage{array}
\usepackage{ragged2e} 
\usepackage{makecell} 
\usepackage{fontawesome5}
\definecolor{mapuche}{RGB}{226, 163, 78}
\definecolor{tupiguarani}{RGB}{154, 117, 222}
\definecolor{guaycuru}{RGB}{51, 223, 195}
\definecolor{quechua}{RGB}{177, 217, 96}
\definecolor{mataco}{RGB}{23, 202, 39}
\definecolor{aymara}{RGB}{255,255,255}
\definecolor{chon}{RGB}{220,68,212}

\usepackage{tikz}
\usepackage{natbib}
\usepackage{hyperref}

\newcommand{\tikzcircle}[2][red,fill=\guaycuruan]{\tikz[baseline=-0.7ex]\draw[#1,radius=#2] (0,0) circle ;}%

\title{
Indigenous Languages Spoken in Argentina: \\A Survey of NLP and Speech Resources}

\author{
  \textbf{Belu Ticona\textsuperscript{1,2}},
  \textbf{Fernando Carranza\textsuperscript{3,4}},
  \textbf{Viviana Cotik\textsuperscript{2,5}}
\\
  \textsuperscript{1}{George Mason University, United States}\\ 
  \textsuperscript{2}{Departamento de Computación, FCEyN, Universidad de Buenos Aires (UBA), Argentina} \\
  \textsuperscript{3}{Departamento de Letras, FFyL, UBA, Argentina}\\
  \textsuperscript{4}{Instituto de Filología y Literaturas Hispánicas ``Dr. Amado Alonso'', UBA, Argentina}\\
  \textsuperscript{5}{Instituto de Investigación en Ciencias de la Computación (ICC), CONICET-UBA, Argentina}\\
  \small{
    \textbf{Correspondence:} \href{mailto:email@domain}{mticona@dc.uba.ar}
  }
}

\begin{document}
\maketitle

\begin{abstract}
Argentina has a large yet little-known Indigenous linguistic diversity, encompassing at least 40 different languages. The majority of these languages are at risk of disappearing, resulting in a significant loss of world heritage and cultural knowledge. Currently, unified information on speakers and computational tools is lacking for these languages. In this work, we present a systematization of the Indigenous languages spoken in Argentina, classifying them into seven language families: Mapuche, Tupí-Guaraní, Guaycurú, Quechua, Mataco-Mataguaya, Aymara, and Chon. For each one, we present an estimation of the national Indigenous population size, based on the most recent Argentinian census. We discuss potential reasons why the census questionnaire design may underestimate the actual number of speakers. We also provide a concise survey of computational resources available for these languages, whether or not they were specifically developed for Argentinian varieties.


\end{abstract}

\section{Introduction}
\defcitealias{INDEC2024}{INDEC, 2024}
\defcitealias{thelangasproject}{The Langas Project, 2012}

By the end of this century, about half of all languages spoken in the world are in danger of disappearing, according to UNESCO \citep{unesco2010}. Since language is a key part of the identity and culture of speakers, the development of technology may help sustain and promote linguistic diversity and maintain cultural heritage. 

Developing technology for endangered languages has the potential to help communities preserve and revitalize their cultural and linguistic heritage, enhance digital communication, increase access to information, and improve education in their native languages, among other benefits. 
In this context, the natural language processing (NLP) community has been putting efforts into developing computational resources for languages around the world, including Indigenous languages from Latin America \cite{tonja2024nlp}, and under-served languages for specific countries and regions \cite{Indonesiaaji2022one,Italyramponi2024language, Africagotowards, Germanicblaschke2023survey}.

However, technical and ethical challenges emerge in adapting common NLP practices and techniques when working with Indigenous languages and their speaker communities \citep{mager-etal-2018-challenges, birdDecolonisingSpeechLanguage2020, liuNotAlwaysYou2022, schwartzPrimumNonNocere2022,  magerEthicalConsiderationsMachine2023, bird-2024-must}. For instance, it is crucial to ensure that these technologies align with the specific needs and priorities of the communities they aim to support. 

In Argentina, according to the National Institute of Statistics and Censuses \citepalias{INDEC2024}, there are 58 Indigenous Peoples, 40 Indigenous languages, and approximately 1.3 million Indigenous descendants, of which only 29.3\% are speakers of an Indigenous language. In contrast to other countries in the region, the establishment of a narrative of European descendance has historically shaped the sociopolitical and cultural agenda, marginalizing the Indigenous population and creating a lack of social awareness regarding the cultural diversity of the country \citep{quijadaNationalMyths2004,adamovskyColorNacionArgentina2012}. 

This is evident in the fact that there is only one published linguistic survey of Indigenous languages in Argentina \cite{censabellaLenguasIndigenasArgentina1999}\footnote{Briefer, not specific, partial or unpublished work include \citet{ciccone2010}, \citet{Censabella2009chaco} and \citet{{Nercesian2021}}, among others.}. The lack of enough information on Indigenous languages, and 
the lack of reliable data on the number of speakers and sociolinguistic situation complicates the assessment of the state of computational resources available for these languages. 
To address this issue, we explore the status of Indigenous languages spoken in Argentina focusing on their prevalence, number of speakers, and available NLP resources, along with a discussion of the main trends that characterize them. This work could serve as a valuable resource for new research groups, helping them to quickly familiarize themselves with key topics, tools, and ongoing debates in the field. 

\noindent We contribute by providing:
\begin{enumerate} [topsep=6pt]
    \setlength{\itemsep}{2pt}
    \setlength{\parskip}{0pt}
    \setlength{\parsep}{0pt}
    \item An overview of the linguistic diversity of Indigenous languages in Argentina,
    \item A survey of computational resources and regional work done for these languages.
\end{enumerate}

The rest of the paper is organized as follows. In Section \ref{section-language-diversity} we present an overview of the Indigenous languages spoken in Argentina. In Section \ref{section-NLPindigenouslanguages}, we review the work done in the past for these languages. In Section \ref{section-challenges}, we discuss the general trends derived from our survey. In Sections \ref{conclusions} and   \ref{limitations}, we provide conclusions 
and limitations of our work. In the \hyperref[sec:appendix]{Appendix}, we explain the methodology employed to collect and select the papers, we provide additional information about the data sources used to create Figure \ref{fig:figLenguas}, and two tables providing an overview of the available corpora and tasks studied for the Indigenous language families Mapuche, Tupí-Guaraní, Quechua, and Aymara.

\section{Indigenous Languages in Argentina}
\label{section-language-diversity}
Currently, there is no consensus on the precise list of languages spoken in Argentina, as demonstrated by comparison of sources such as \citet{censabellaLenguasIndigenasArgentina1999, Censabella2009chaco}, \citet{ciccone2010} and \citet{Nercesian2021}. 
This uncertainty is due to the complex situation of each language and its speakers, which covers language endangerment, as well as different situations regarding the amount of available documentation, standardization, use contexts, and the availability of a written form, among others.
Furthermore, speakers of some of these languages often exhibit a negative attitude toward their linguistic heritage, complicating its study \citep{carrio2014}.

\begin{figure*}[t]
\centering
\begin{minipage}{0.24\linewidth}
\centering
\includegraphics[trim={0.3cm 0.5cm -0.5cm 0.5cm}, clip, width=1.23\textwidth]{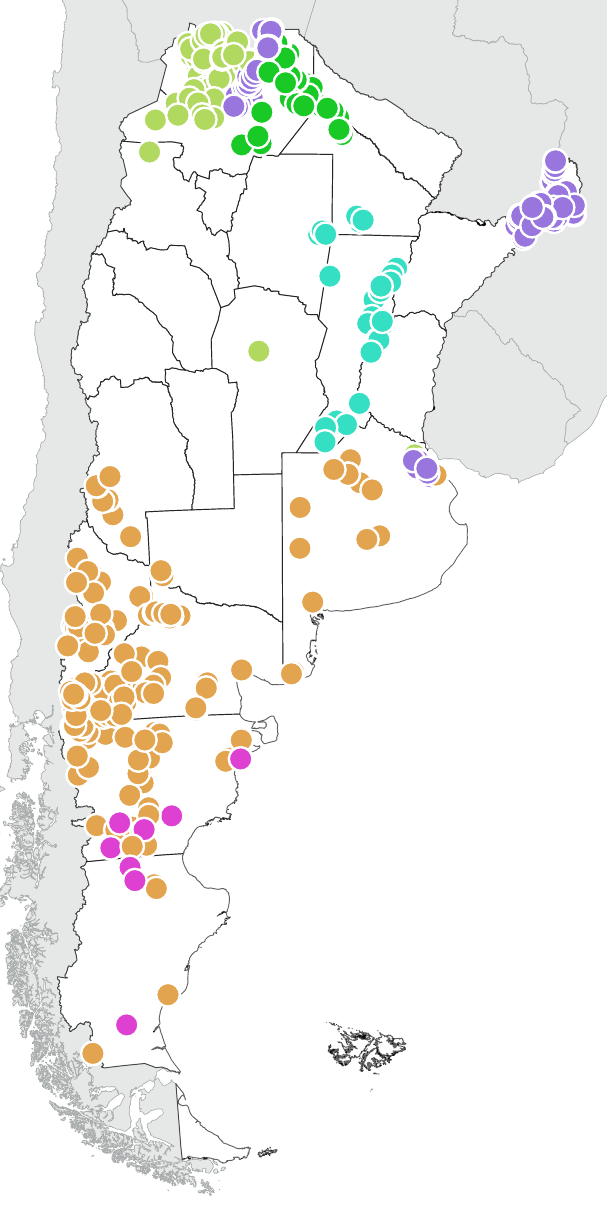}
\end{minipage}
\begin{minipage}{0.75\linewidth}
\centering
\scalebox{0.62}{

    \begin{tabular}{lccccc}
    \toprule
    \multicolumn{6}{l}{\textbf{Language Family}} \\
    Language &\textbf{ISO-639-3} & \textbf{Population} & \textbf{Speaker (\%)} & \textbf{Speaking Population} & \textbf{Vitality} \\
    \midrule
    \textbf{Mapuche} \tikzcircle[black, fill=mapuche]{5pt} \\
    Mapudungún & arn & 145,783 & 18 & 26,240 & {\color{red} \faExclamationCircle} \\
    \midrule
    \textbf{Tupí-Guaraní} \tikzcircle[black, fill=tupiguarani]{5pt} \\
    Ava Guaraní (Chiriguano) & nhd & 3306 & 45 & 1490 & {\color{red} \faExclamationCircle} \\
    Paraguayan Guaraní & grn-gug & \multirow{2}{*}{\hfil 135,232} & \multirow{2}{*}{\hfil 40} & \multirow{2}{*}{\hfil 54,092} &  {\color{blue} \faInfoCircle} \\
    Corrientes Guaraní & grn &  &  & & {\color{Green} \faShield*} \\
    Mbyá Guaraní & gun & 11,014 & 84 & 9,251 & {\color{Green} \faShield*} \\
    Tapiete & tpj & 654 & 39 & 255 & {\color{red} \faExclamationCircle} \\
    \midrule
    \textbf{Guaycurú} \tikzcircle[black, fill=guaycuru]{5pt} \\
    Toba (Qom) & tob & 80,124 & 49 & 39,260 & {\color{red} \faExclamationCircle} \\
    Mocoví & moc & 18,231 & 24 & 4,375 & {\color{red} \faExclamationCircle} \\
    Pilagá & plg & 6,169 & 89 & 5,489 & {\color{red} \faExclamationCircle} \\
    \midrule
     \textbf{Quechua} \tikzcircle[black, fill=quechua]{5pt} \\
    Kolla Quechua & que & 69,121 & 41 & 28,339 & No data \\
    Cusco-Bolivian Quechua & que-quz & \multirow{2}{*}{\hfil 52,154} & \multirow{2}{*}{\hfil 51} & \multirow{2}{*}{\hfil 26,598} & {\color{blue} \faInfoCircle} \\
    Santiago del Estero Quichua & que-qus &  &  &  & {\color{Green} \faShield*} \\
    \midrule
     \textbf{Mataco-Mataguaya} \tikzcircle[black, fill=mataco]{5pt}\\
    Wichí & wlv/mzh & 69,080 & 73 & 50,428 & {\color{Green} \faShield*} \\
    Chorote & crq/crt & 3,238 & 75 & 2,428 & {\color{Green} \faShield*}/{\color{red} \faExclamationCircle} \\
    Nivaclé & cag & 878 & 75 & 658 & {\color{Green} \faShield*} \\
    \midrule
     \textbf{Aymara} \tikzcircle[black, fill=aymara]{5pt} \\
    Central Aymara & aym-ayr & 19,247 & 51 & 9,815 & {\color{Green} \faShield*} \\
    \midrule
    \textbf{Chon} \tikzcircle[black, fill=chon]{5pt} \\
    Tehuelche & teh & 17,420 & 8 & 1,393 & {\color{red} \faExclamationCircle} \\
    \bottomrule
    \end{tabular}

}
\end{minipage}
\caption{Geographical and demographic description of Indigenous languages spoken in Argentina by more than 1000 speakers \citepalias{INDEC2024}. To the left, the distribution of the Indigenous communities homonymous to Indigenous languages. To the right, Indigenous languages spoken in Argentina, grouped by family and identified by their ISO 639-3 code (macrolanguage identifier, and microlanguage in cases when more than one variety is mentioned). For each case, the population and the corresponding percentage of speakers are provided, with the percentage reflecting the proportion of the community that speaks the Indigenous language they consider to be the language of their community, as reported in the Argentine INDEC census. All data was obtained from the national census \citepalias{INDEC2024}. \emph{Speaking population} shows the estimated number of speakers of the language based on the Indigenous Population and the previously described percentage. \emph{Vitality} shows the level of endangerment of the language according to Ethnologue \citep{eberhard2015ethnologue}: {\color{Green} \faShield*}, {\color{red} \faExclamationCircle} and {\color{blue} \faInfoCircle} denote stability, endangerment and institutional status of the language, respectively. When there is no available data to distinguish between varieties associated with a given language, the data is placed in the middle of the lines. Aymara is not shown in the map, because there is no community data from Aymara in the INDEC data.}
  \label{fig:figLenguas}
\vspace{-14pt}
\end{figure*}

Based on the works of \citet{censabellaLenguasIndigenasArgentina1999} and \citet{ciccone2010}\footnote{See also \citet{Nercesian2021}.}, we consider the following language families: Mapuche, Tupí-Guaraní, Guaycurú, Quechua, Mataco-Mataguaya, Aymara, and Chon. When relevant, we also include notes on the peculiarities of Argentine varieties. In Figure \ref{fig:figLenguas}, we present an overview of the Indigenous language families reviewed in this section along with their demographic data and geographical location specific to Argentina. More information about other languages not included in this survey and  resources used to build the figure is available in \hyperref[sec:appendix]{Appendix}. 

It is important to note that the available data is not sufficiently reliable due to a methodological issue. In the 2022 National Census, only individuals who self-identified as members of an Indigenous community were asked whether they spoke the Indigenous language of their community, without the census inquiring which specific language it was. Thus, speakers of Indigenous languages who do not self-identified as members of an Indigenous community or who speak a language of another community were not considered. This non-self identification might be due to historical discrimination, and the socially negative connotation of \textit{being indigenous} as opposed to \textit{being Argentinian}\footnote{As \citet{quijadaNationalMyths2004} explains, the construction of Argentinian identity as a 'white nation', homogeneously integrated with European roots, was built on myths that deny any other possible identity. One of the those myths is the collective belief that 'there is no indigenous (\textit{indios}) since they (first governors) killed them all'.}. On top of this, even though in theory all family members are supposed to be asked, in practice only a family member is generally asked by census facilitators. 

\noindent From the \textbf{Mapuche} family, Mapudungún or Mapuzungún\footnote{See \citet{diaz2006glosonimos} for details on the Glossonyms for this language family.} is the most spoken Indigenous language in Argentina and Chile. According to \citet{ViegasBarros:1999}, the varieties spoken in both countries differ 
 mainly in pronunciation and vocabulary.

\noindent The \textbf{Tupí-Guaraní} family is primarily spoken in the south of the Amazon. In Argentina, it comprises Ava Guaraní (also known as Chiriguano), Tapiete 
(also considered an Ava Guaraní variety \citep{Dietrich1986chiriguano}), Mbyá Guaraní and two varieties of Guaraní: Corrientes Guaraní, the Argentinian variety spoken in the Corrientes province, and Paraguayan Guaraní, primarily spoken by Paraguayan immigrants specially in Buenos Aires, Misiones and Formosa provinces \cite{ciccone2010}. The peculiarities of the Corrientes variety, as compared with the Paraguayan one, seem to lie in the amount of Spanish loanwords, the pronunciation of some phonemes, the use of \textit{ta} instead of \textit{piko} as an interrogative morpheme, its set of evidential particles and the use of \textit{m\~a} as intensifier \citep{cerno2010evidencias,cerno2013guarani}.

\noindent The \textbf{Guaycurú} family includes Toba (or Qom, as their speakers call it), Mocoví, and Pilagá. These languages are spoken in the north of Argentina, mainly in the provinces of Formosa and Chaco.

\noindent The \textbf{Quechua} family covers different varieties spoken in Argentina, Bolivia, Colombia, Chile, Ecuador and Perú (\citealt[p. 22]{ATLASUNICEF2009}). This makes it the most extended Indigenous language of South America, both in number of speakers (over 6 millions according to \citealt[p. 101]{unesco2010}) and in countries where it is spoken (6 countries, \citealt[p. 76]{ATLASUNICEF2009}). In Argentina, two Quechua varieties are identified: Cusco-Bolivian Quechua primarily spoken by Bolivian and Peruvian immigrants, and Santiago del Estero Quichua, which is mostly spoken by the local population of Santiago del Estero, a province in Argentina (\citealt[p. 24]{Juanatey2020tesis}).  
According to \citet{Adelaar1995raices} and \citet{Juanatey2020tesis}, the particularities of the Santiago del Estero Quichua include the loss of the proto-quechua semivowel /w/ between vowels, the changes in plural morphology and verbal inflection, and some specific lexical choices. 
In some literature, a Kolla Quechua variety is included (e.g., \citealt{censabellaLenguasIndigenasArgentina1999}). Kollas are an heterogeneous Indigenous community. We are not aware of any work addressing whether the Kolla Quechua shows peculiarities that justify treating it as a different variety.

\noindent The \textbf{Mataco-Mataguaya} family is spoken in Paraguay and Argentina. In Argentina, it includes the Wichí language, spoken in Formosa, the Nivaclé or Churupí, spoken in Salta and Formosa, and Chorote, spoken in Salta.

\noindent The \textbf{Aymara} family comprises different Aymara varieties. In Argentina, the most extended is considered to be the Central Aymara variety \citep{ciccone2010}. We are not aware of any work addressing specifically the peculiarities of the Central Aymara variety spoken in Argentina. 

\noindent Finally, the \textbf{Chon} family consists of several languages that were spoken in Patagonia. The majority are now extinct or in severe danger. One of the surviving languages (according to current knowledge, it might be the last) is Tehuelche, which is spoken in the Santa Cruz province \cite{censabellaLenguasIndigenasArgentina1999}).

\section{Survey of Computational Resources}
\label{section-NLPindigenouslanguages}
\vspace{-0.05cm}
In this section, we present our survey of NLP research related to the Indigenous language families spoken in Argentina. 
It is worth noting that most of the resources we found were not developed specifically for Argentine varieties. 

Due to the lack of resources for Argentine varieties, and for the sake of completeness, we decided to include research on varieties spoken not only in Argentina but also in other countries. Therefore, some language varieties mentioned in this section may differ from those listed in Section \ref{section-NLPindigenouslanguages}. 
The protocol followed, to collect and select the papers from which the available resources and tasks are outlined, is described in the section \hyperref[subsec:AppProtocol]{Employed Protocol} of the Appendix. Tables \ref{survey_resources} and \ref{survey_tasks} (see \hyperref[subsec:TablasCorporaYTasks]{Appendix})  summarize the corpora and tasks found for each of the considered language families.

\vspace{-0.2cm}
\subsection{Mapuche Family}

Despite being the language of the largest Indigenous Peoples in Argentina and Chile, there is only one public large corpus for Mapundungún language \cite{duanResourceComputationalExperiments2020}. 
This corpus is a clean and detailed version of the recordings collected during the AVENUE project \citep{levin2000data}. 

As a result of this collaboration, rule-based MT systems were developed for Mapudungún-Spanish, as well as a spelling checker for Mapudungún \cite{monson2004data,monson2008linguistic,llitjos2005building, monson2006building}. 
None of these resources are publicly available anymore. Related work can also be found in \citet{pendasNeuralMachineTranslation2023}. Regarding the educational perspective, \citet{ahumada2022educational} designed some tools for educational purposes, including an orthography detector and converter, a morphological analyzer, and an informal translator.

\subsection{Tupí-Guaraní Family}

Among the Tupí-Guaraní languages, Paraguayan-Guaraní has been consistently covered over time by different NLP researchers, resulting in multiple monolingual corpora and even a parallel corpus in Guaraní-Spanish. However, much work remains to be done for other languages of the family.  
The first computational research initiatives in Guaraní were developed as isolated projects in different groups and countries. These include the work on data collection for sentiment analysis \cite{riosSentimentCategorizationCreole2014, aguero-toralesLogisticalDifficultiesFindings2021}, database collection of historical texts \cite{cordovaProcessingQuechuaGuarani2019,thelangasproject}, and the adaptation of Universal Dependencies guidelines for annotating Mbyá Guaraní \cite{thomasUniversalDependenciesMbya2019a}. 

In recent years, a research collaboration among multiple groups and countries gave birth to Jojajovai, the first medium-sized corpus of the language family \cite{chiruzzoJojajovaiParallelGuaraniSpanish2022}. Previous works detail the challenges of developing a Guaraní corpus, suggesting ideas to diversify the type of content \citep{chiruzzoDevelopmentGuaraniSpanish2020a, gongoraExperimentsGuaraniCorpus2021}. \citet{gongoraCanWeUse2022} used the Jojajovai dataset to enrich MT systems with pre-trained word embeddings. \citet{toralesMachineLearningApproaches2022} 
conducted exhaustive research studying topic modeling and sentiment analysis on text from social media in Guaraní and Jopará, a mixture of Guaraní and Spanish used in Paraguay. For information on Jopará, see \citet{estigarribia2015guarani}.

Recently, the research community has started to work on other Tupian languages\footnote{Tupí is a language family native to South America, that includes various languages spoken primarily in Brazil. Tupí-Guaraní is one of its major subfamilies.}.
For instance, the TuLar project\footnote{\url{https://tular.clld.org/}} collects, documents, and develops computational and pedagogical materials for Indigenous communities in Brazil. \citet{martinrodriguezTupianLanguageResources2022} released an online lexical database with more than 400 concepts, a morphological database with 51 languages, and a dependency treebank with 9 languages.

\subsection{Quechua Family}

The Quechuan languages are the most studied by the NLP community, mainly performed by Peruvian researchers. 

There are projects for creating speech corpora and monolingual text corpora, which cover only Cusco-Bolivian Quechua varieties \citep{cardenasSiminchikSpeechCorpus,zevallosHuqariqMultilingualSpeech2022,paccotacya2022speech, zevallosIntroducingQuBERTLarge2022}. There are also considerable studies conducted on these varieties, which cover common NLP tasks, such as language identification \cite{Linares2017},  
machine translation \citep{ortegaNeuralMachineTranslation2020, Oncevay2021peru, alvarez2023model}, corpora alignment \cite{ortegaUsingMorphemesAgglutinative2018}, lexical database construction \cite{melgarejoWordNetQUDevelopmentLexical2022}, and the creation of resources, such as data augmentation \cite{zevallos2022data}. 

Other efforts have been made in evaluating and applying linguistic tools for Quechua languages, such as a morphological analyzer \cite{himoroPreliminaryResultsEvaluation2022}, and the use of an automatic grammar generator for the study of gerunds in Quechua and Spanish \cite{rodrigoApproachAutomaticTreatment2021}. 
Only one resource specifically created for the Santiago del Estero Quichua was found in our survey: \citet{Porta2010}. 
This study presents a transducer that aims at identifying the morphological structure of the language. 
\vspace{-0.5em}
\subsection{Other families}

For the rest of the families considered in this paper, we found few limited resources. From the Guaycurú Family, we only found a work on spoken language identification for Qom \citep{garber2022sistema} and a description of its morphology using a linear context-free grammar (\citealt{porta2010use}). No special resources were found for Pilagá or Mocoví. To the best of our knowledge, no specific resources were developed for the Mataco-Mataguaya family, besides the fact that Nivaclé was taken into account in the language identification model presented in \citet{kargaran2023glotlid}. 

Regarding Aymara, the few existing works relied on data available through the shared task of the AmericasNLP workshop, exploring Spanish-Aymara machine translation \citep{gillinFewshotSpanishAymaraMachine2023, oncevayPeruMultilingualIts2021}. Finally, regarding the Chon Family, \citet{Domingo2018corpustehuelche} present a publicly available corpus on Techuelche, the only computational resource we found for this family.

\vspace{-0.1em}
\section{Discussion: Trends and Challenges}
 \label{section-challenges}

\textbf{The scarce resources available were produced in Argentina's neighboring countries.}
In this survey, we identified only a few research groups working steadily over time on Indigenous languages spoken in Argentina and its surroundings. Among the handful exceptions, we highlight the Peruvian academic community which has developed most of the work done for the Quechua language family. For other families, most of the available work has been done by a unique research group (e.g. NLP Group of UdelaR\footnote{Universidad de la República, Uruguay.}, in the case of Chiruzzo's work for the Guaraní family in Uruguay) or a particular initiative (e.g. the AVENUE301 project for Mapudugún). Besides these cases, most of the work identified has been conducted primarily by South American researchers working in the diaspora. It is worth mentioning that, in general, research groups from South America have limited access to computing resources and funding. 
\vspace{0.12em}

\noindent \textbf{Local languages and variants have yet to be incorporated into emerging academic initiatives.} A lot of work has been done for the AmericasNLP workshop on Indigenous languages (see 2020-2024 proceedings). This shows the positive impact of these challenges on the field. Nevertheless, there is almost no work conducted on the Argentinian local varieties. 
 As seen in Section \ref{section-language-diversity}, the peculiarities of Argentinian varieties are scarcely studied for Corrientes Guaraní and Santiago del Estero Quichua, less known for Mapudungún and almost ignored for Aymara. 
 For this reason, it is difficult to assess how effectively resources developed in other countries might work for local varieties.
\vspace{0.12em}

\noindent \textbf{More focus on written languages, while indigenous languages are traditionally oral.}
Additionally, we found that the academic community tends to highlight technical challenges encountered when adopting approaches commonly used for standardized languages. 
Most of the surveyed work uses techniques developed for written languages, while most Indigenous Peoples use their languages predominantly in a spoken form. Since modern approaches rely on data availability in written forms and computing power, the technical challenges are typically framed from the perspective of data scarcity (e.g. lack of parallel data, and lack of orthographic normalization, among others).
\vspace{0.1em}

\noindent \textbf{Inclusion of Indigenous People.} Finally, we would like to point out that only a few exceptions consider an evaluation based on the needs and use of Indigenous Peoples. For example, \citet{ahumada2022educational} provide detailed feedback given by the Indigenous descendants, and an in-depth case study on usability. Inclusion of native voices could leverage the applications and impact of the resources released by the academic community.

\section{Conclusions}
 \label{conclusions}
   In this paper, we survey computational resources for the most spoken Indigenous languages in Argentina. To better comprehend their language diversity, we present an overview of demographic data for the Indigenous Peoples most present in the country. Among the seven language families considered, we find that most NLP applications and resources are developed for the Quechua, Tupí-Guaraní, and Mapuche families, often in varieties different from those spoken in Argentina. 

\section{Limitations}
   \label{limitations}
   The authors of this work are culturally situated in academic contexts of hard access for indigenous identities in Argentina. Only one of us self-identifies as an indigenous descendant. We acknowledge that to work in this area an interdisciplinary approach is needed with members of the Indigenous communities being part of it.

\section*{Acknowledgments}
   \label{acknowledge}
   Belu Ticona is partially funded by the generous support of the US National Science Foundation under grants CNS-2234895 and IIS-2327143.

\bibliography{coling_latex.bib}

\appendix

\section*{Appendix: Data Sources, Figures and Tables}
\label{sec:appendix}
In this section we provide additional information about the methodology employed to collect and select the papers, about the 
data sources used to create Figure \ref{fig:figLenguas} (in Section \ref{section-language-diversity}), and also two tables providing an overview of the available corpora and tasks studied for the Indigenous language families Mapuche, Tupí-Guaraní, Quechua, and Aymara.

\subsection*{A. Employed Protocol}\label{subsec:AppProtocol}

In order to collect and select the papers from which the available resources and tasks were outlined, we gathered information from the Proceedings of the most relevant venues in the field: the Workshop on Natural Language Processing for Indigenous Languages of the Americas (AmericasNLP), the Association for Computational Linguistics (ACL) (main conference and workshops -such as Use of Computational Methods in the Study of Endangered Languages -ComputEL-, Workshop on Technologies for MT of Low Resource Languages (LoResMT), and Workshop on Deep Learning Approaches for Low-Resource NLP-),  Language Resources and Evaluation Conference (LREC), Conference on
Computational Linguistics (COLING), Conference of the North
American Chapter of the Association for Computational Linguistics (NAACL), and papers referred by selected papers from these sources. 

\subsection*{B. Data Sources}\label{subsec:DataSources}

The right part of Figure \ref{fig:figLenguas}, was created as follows. Language families and languages were based on those described by the literature \citep{Adelaar:2012,Adelaar:2010south,Censabella2009chaco,censabellaLenguasIndigenasArgentina1999,Nercesian2021,ciccone2010}. Those languages were mapped to Ethnologue \citep{eberhard2015ethnologue}\footnote{\url{https://www.ethnologue.com/}}, from where the ISO 639-3 code (i.e. macrolanguage identifier) was obtained. In cases when more than one variety is mentioned, the microlanguage identifier is provided in the \emph{ISO 639-3} column. Data regarding Indigenous population and \% of speakers was obtained from the Argentine National Census Data \cite{INDEC2024}\footnote{The results were taken from \url{https://censo.gob.ar/index.php/datos_definitivos_total_pais/}, especially from tables 8 and 9.}. 

These numbers only reflect the number of people self-identifying as part of an Indigenous community and, among them, the number of people who consider themselves to speak the Indigenous language of that community. For this reason, the data on speaking population may not fully represent the number of Indigenous language speakers: some communities have lost their ancestral languages and speak other Indigenous languages, while individuals outside the Indigenous population may still speak Indigenous languages (\citealt{ciccone2010}, \citealt[pg.~159-169]{Censabella2009chaco}). The estimated number of speakers is based on the data from the previous two columns. Finally, vitality, the level of endangerment of the language, was completed according to Ethnologue web page. 

It is important to mention that not all information about Indigenous Peoples' communities provided by the national census, can be mapped to ISO languages straightforwardly. For instance, there is no special ISO code for the variety spoken by the Kolla community, considered in the census. Inversely, there is no Aymara community, among the communities considered in the census, which does not imply there is no speaker of this language, which is included in the ISO codes. 

In order to calculate number of speakers from the INDEC data, we only used as reference those Indigenous communities whose names are identical to the language name, except for the number of speakers of the Ava Guaraní (Chiriguano), where we summed, following \citet{ciccone2010}, the number of people who self-identified as member of Chané and Isoceño groups. 
For some languages, there is no specific reference community. For instance, there is no distinction among Paraguayan Guaraní and Corrientes Guaraní or among Santiago del Estero Quichua and Cusco-Bolivian Quechua). In those cases, we vertically center aligned the available data (population, speaker ratio and speaking population).

\defcitealias{CUI2024}{}

The geographical distribution of the indigenous communities (left part of Figure \ref{fig:figLenguas}) was obtained from the data from the 2022 national census \citepalias{INDEC2024}\footnote{See \url{https://datos.jus.gob.ar/dataset/listado-de-comunidades-indigenas/archivo/8f9af332-83ff-4ea6-a6a5-7a90371a41fb}}. We used QGIS\footnote{\url{https://www.qgis.org}} to plot the location of the communities. Other publicly available maps consulted for this work include the map done by the University Language Center (CUI)\footnote{https://cui.edu.ar/pdf/MapadelasLenguasIndigenasenArgentina2024-08.pdf}, which also shows the provincial distribution of languages in Argentina using census data, statistical visualizations of census data provided by \citepalias{INDEC2024} on their platform, and maps published by the National Institute of Indigenous Affairs (INAI) \footnote{https://www.argentina.gob.ar/derechoshumanos/inai/mapa}.     

Information on the number of languages speakers in different countries can also be obtained from UNESCO's World Atlas of Languages (WAL) \citep{unesco2010}\footnote{\url{https://en.wal.unesco.org}}. Nevertheless, we showed data from the national census (INDEC), which is the primary source for Argentina (and that differs from the information provided by UNESCO's WAL).  
The status of languages regarding their danger of disappearing can also be found in Glottolog \citep{glottologharald_hammarstrom_2023_8131084} -which also shows all the varieties of the language families- and UNESCO's WAL. 
Finally, additional (and different) information regarding Indigenous languages spoken in Argentina can be seen in the Observatorio de los Derechos de los pueblos Indígenas y Campesinos\footnote{\url{https://www.soc.unicen.edu.ar/observatorio/index.php/22-articulos/106-unas-700-000-personas-mantienen-vivas-15-lenguas-indigenas-en-argentina}}.
F

\subsection*{C. Available Corpora and Tasks}\label{subsec:TablasCorporaYTasks}

Next, we present two tables (Tables \ref{survey_resources} and \ref{survey_tasks}), that provide an overview of the available corpora and tasks studied for the Indigenous language families Mapuche, Tupí-Guaraní, Quechua, and Aymara.

\renewcommand{\arraystretch}{1.25}
\begin{table*}[t]
\footnotesize
\centering
\begin{tabular}
{
p{.055\textwidth}
p{.1248\textwidth} 
>{\centering}p{0.02\textwidth}  
p{.17\textwidth} 
p{.29\textwidth} 
p{.18\textwidth}l
}
\hline
\textbf{Family} & \textbf{Variety} & \textbf{Area} & \textbf{Task} & \textbf{Size \& Description} & \textbf{Paper} \\
\hline
Mapuche & Mapudungún & T & machine translation & 384k sentences, medical domain & \citet{pendasNeuralMachineTranslation2023} \\
& Mapudungún & S & speech recognition, speech synthesis and
machine translation & 142 hs, transcribed audio, medical domain  & \citet{duanResourceComputationalExperiments2020} \\
\hline
 Tupí-Guaraní  & Paraguayan Guaraní
 & T & machine translation & 30k sentences, parallel Spanish-Guaraní data collected from news, folktales, articles, biographies (*)  & \citet{chiruzzoJojajovaiParallelGuaraniSpanish2022, chiruzzoDevelopmentGuaraniSpanish2020a} \\
  & Multiple varieties & T & multiple tasks & dependency treebanks, morphological and lexical datasets& \citet{martinrodriguezTupianLanguageResources2022} \\
  \hline
Quechua   & South & T & machine translation & 127k sentences, Spanish-Quechua parallel data, legal domain, biblical domain (*) &  \citet{degibertFourApproachesLowResource2023,ebrahimiFindingsAmericasNLP20232023,ahmed-etal-2023-enhancing, agicJW300WideCoverageParallel2019, tiedemannParallelDataTools2012} \\
& Chanca, Collao & T & NER, POS tagging& 384k sentences,  multiple domains as religion, education, health, narrative, social (*)& \citet{zevallosIntroducingQuBERTLarge2022} \\
  & South., Central, \par North., Amazon & T & POS tagging & 29k words, lexical resources for the development of a Quechua \textit{wordnet} (*)  & \citet{melgarejoWordNetQUDevelopmentLexical2022} \\
  & Central, South & S & speech recognition, language identification, text-to-speech & 220 hs, transcribed audio & \citet{zevallosHuqariqMultilingualSpeech2022}\\
 & Collao & S & emotion recognition & 15 hs, raw audio (*)& \citet{paccotacya2022speech}\\
 & Chanca, Collao & S & speech recognition & 97 hs, raw audio from radio shows & \citet{cardenasSiminchikSpeechCorpus}\\
 \hline
Aymara & Unspecified & T &  machine translation & 900 Aymara-English pairs, lexical dataset from dictionaries  & \citet{gillinFewshotSpanishAymaraMachine2023} \\
 & Unspecified & T &  machine translation & 25k sentences aprox., multiples sources (legal, bliblical) (*) & \citet{degibertFourApproachesLowResource2023} \\
  & Central & T &  machine translation & 3k sentences., multiples domains (news, health, informal and formal register) (*) & \citet{teamNoLanguageLeft2022}\\
 & Unspecified & T &  machine translation & 6.5k sentences, multiple domain (*) & \citet{tiedemannParallelDataTools2012} \\

\hline
\end{tabular}
\caption{\label{survey_resources}
Overview of available corpora for the Indigenous language families Mapuche, Tupí-Guaraní, Quechua and Aymara.
 South, North, and Amazon stand for Southern, Northern, and Amazonian, respectively. S and T stand for Speech and Text respectively. An asterisk (*) in the \textit{Size \& Description} column indicates that the resource is publicly available. The table shows research on languages spoken in Argentina, but for varieties spoken in other countries. Therefore, some language varieties mentioned in this section differ from those listed in Section \ref{section-language-diversity}.
 }
\end{table*}

\begin{table*}[t]
\footnotesize
\centering
\begin{tabular}{p{.13\textwidth}p{.2\textwidth}p{.3\textwidth}l}
\hline
\textbf{Family} & \textbf{Task} & \textbf{Paper}\\

\hline
 Mapuche & MT (text) & \citet{pendasNeuralMachineTranslation2023, levin2000data, duanResourceComputationalExperiments2020,monson2004data,monson2008linguistic,llitjos2005building, monson2006building} \\
   & linguistic tools & \citet{chandiaMapudungunFSTMorphological2022} \\
          & educational tools (text) & \citet{ahumada2022educational} \\
  \hline
 Tupí-Guaraní  & language identification (text)  & \citet{cavalinUnderstandingNativeLanguage2023} \\
 & sentiment analysis  & \citet{riosSentimentCategorizationCreole2014, aguero-toralesLogisticalDifficultiesFindings2021} \\
  & MT (text) & \citet{lucas-etal-2024-grammar, pinhanez-etal-2024-human, gongoraCanWeUse2022} \\
 & code switching (text) & \citet{chiruzzoOverviewGUASPAIberLEF2023,jauhiainenTuningHeLIOTSGuaraniSpanish2023, toralesMachineLearningApproaches2022} \\
  & topic modelling (text)& \citet{toralesMachineLearningApproaches2022} \\
  & educational tools (text) & \citet{bui-von-der-wense-2024-jgu, martinrodriguezTupianLanguageResources2022} \\
  \hline
Quechua & NER & \citet{zevallosIntroducingQuBERTLarge2022}\\
 & POS tagging & \citet{zevallosIntroducingQuBERTLarge2022}\\
 & language identification (speech)  & \citet{paccotacya2022speech,Linares2017}\\
  & MT (text) &  \citet{tiedemannParallelDataTools2012, ahmed-etal-2023-enhancing,vazquez2021helsinki} \\
 & emotion recognition  & \citet{paccotacya2022speech}\\
  & speech recognition  & \citet{zevallosHuqariqMultilingualSpeech2022} \\
  & text-to-speech & \citet{zevallosHuqariqMultilingualSpeech2022} \\
  & morphological analysis & \citet{Porta2010,porta2010use}\\
\hline
 Aymara & MT (text) & \citet{gillinFewshotSpanishAymaraMachine2023,alanocaNeuralMachineTranslation2023, oncevayPeruMultilingualIts2021} \\
  & language identification (text) & \citet{Linares2017}\\
  & linguistic tools & \citet{himoroPreliminaryResultsEvaluation2022, beesleyAymara2003} \\
\hline
\end{tabular}
\caption{\label{survey_tasks}
Overview of tasks studied for the Indigenous language families Mapuche, Tupí-Guaraní, Quechua, and Aymara. 
NER, POS, and MT stand for named entity recognition, part of speech, and machine translation respectively. 
}
\end{table*}

\end{document}